\documentclass[10pt,journal]{IEEEtran}
\usepackage[T1]{fontenc}
\usepackage{amsmath,amssymb,amsfonts}
\usepackage{booktabs}
\usepackage{graphicx}
\usepackage{multirow}
\usepackage{xcolor}
\usepackage{url}
\usepackage{cite}

\newcommand{\QB}{Q_B}
\newcommand{\AL}{a_{\mathrm L}}
\newcommand{\AB}{A_B}

\newcommand{\Reg}{R}

\newcommand{\Ntok}{N_{\mathrm{tok}}}
\newcommand{\Ncf}{N_{\mathrm{cf}}}

\begin{document}
\title{Selective Amortization of Full-Budget Counterfactual Reasoning for Visual Token Communication}
\author{Qinglei Qi, Zhihe Liang, Fengzhan Jing, Shenao Zhu, Lei Zhang, \\ Chenyang Zhang, Shuqing He, Jia Guo*, \IEEEmembership{Member,~IEEE}

 \thanks{This work was supported by the National Natural Science Foundation of China under Grant No.~62002263, the Tianjin Science and Technology Program Projects under Grant No.~24YDTPJC00630, and the Tianjin Research Innovation Project for Postgraduate Students under Grant No.~2026KYCX001F.}
\thanks{Q.Qi, Z.Liang are with the School of Artificial Intelligence, Nanyang Normal University, Nanyang, China. ( e-mail: qiqinglei@nynu.edu.cn, liangzhihe329@gmail.com)}

 \thanks{F.Jing, S.Zhu, L.Zhang Y.Zhang and J. Guo are with the School of Computer and Information Engineering, Tianjin Normal University, Tianjin, China (e-mail: 2611090041@stu.tjnu.edu.cn
,zhushenao34@gmail.com,2611090026@stu.tjnu.edu.cn, 2411090038@stu.tjnu.edu.cn, c04s316@bupt.cn}
 \thanks{S. He is with the School of Information Science and Engineering, Linyi University, Linyi 276000, China (email:heshuqing@lyu.edu.cn)}
 \thanks{*Corresponding Author: Jia Guo} 
  \thanks{The source code for this work is publicly available at: https://github.com/c04s316/-ACV-Gate-}
  }

\maketitle

\begin{abstract}
Generative image communication transmits compact semantic tokens under a limited packet budget, where token selection directly affects the final reconstruction quality after the complete packet is decoded. However, accurately estimating the terminal value of every candidate token requires repeated receiver-side reconstruction, resulting in substantial encoder-side computation. To address this problem, we propose ACV-Gate, an adaptive candidate evaluation framework that learns to approximate full-budget counterfactual evaluation and selectively assigns exact evaluations to the most informative candidates. Specifically, a set-aware student is trained using terminal advantages and regrets to predict candidate rankings directly, while a selective refinement mechanism evaluates only a bounded candidate set containing both Local-MDL and direct actions; cost-based thresholds further enable explicit control of the average evaluation workload. Experiments on CIFAR-10 show that ACV-Gate consistently improves reconstruction quality while substantially reducing candidate evaluations; at 0.20~bpp, the primary adaptive configuration improves PSNR over Local-MDL by $0.636\pm0.119$~dB with only $2.13\pm0.53$ candidate evaluations per image, corresponding to 27.60\% of the calls required by the Exact-Full expert. Matched-candidate comparisons, synchronized GPU measurements, and evaluations on STL-10 and $384\times384$ scale transfer further demonstrate consistent quality--computation trade-offs, with particularly pronounced gains at low bit rates. These results show that combining terminal-value learning with selective candidate evaluation provides an effective and controllable mechanism for allocating encoder computation in packet-constrained generative image communication.

\end{abstract}


\section{Introduction}
Bandwidth-constrained visual communication is important for applications in which images must be delivered under limited or time-varying transmission resources, such as wireless links, edge visual systems, and progressive image delivery. Generative receivers provide a promising mechanism for such settings by reconstructing missing visual content from transmitted information together with learned image priors. Instead of describing every image component explicitly, the transmitter can allocate its packet budget to a subset of informative visual tokens and allow the receiver to complete the remaining content. This communication paradigm makes the selection of transmitted tokens a central design problem: under a fixed packet budget, the transmitted subset should provide the greatest improvement in the final reconstructed image.

The value of an individual token, however, is determined by more than its local image content. It depends on the tokens that have already been transmitted, the tokens that will be selected subsequently, and the way in which the generative receiver completes the untransmitted content. Consequently, effective token selection requires an estimate of the reconstruction quality obtained after the available transmission budget has been fully used. This terminal perspective is particularly important at low bit rates, where each transmitted token occupies a substantial fraction of the packet budget and its interaction with the receiver prior can strongly influence the final reconstruction.

Learned image compression has established a direct connection between rate allocation and reconstruction quality \cite{balle2016endtoend,balle2018hyperprior,minnen2018joint}, while discrete representations and masked generation provide a natural token-based interface for image representation and completion \cite{oord2017vqvae,esser2021taming,chang2022maskgit}. Deep joint source--channel coding and semantic transmission further exploit learned representations and receiver-side priors to improve visual communication under constrained channels \cite{bourtsoulatze2019deep,xie2021deep,han2022generative}. Progressive, content-adaptive, masked, and sparse transmission methods extend these ideas to variable communication conditions by allocating transmission resources according to image content and communication requirements \cite{plit2025,catjscc2025,sqgan2025,sparsesbc2025}. Together, these developments provide increasingly flexible mechanisms for representing and transmitting visual information. For a fixed tokenizer and generative receiver, an important remaining question is how to identify the token positions whose transmission contributes most to the final reconstruction under the available packet budget.

Existing token-pruning and selection methods commonly estimate token importance from attention, saliency, diversity, or coverage \cite{rao2021dynamicvit,liang2022evit,ryoo2021tokenlearner,yin2022avit,bolya2023tome,divprune2025,vispruner2025,scope2025}. Learned selectors and adaptive compression methods further reduce visual computation by predicting which tokens or features deserve additional processing \cite{sparsevlm2025,prumerge2025,fitprune2025,gprune2025,tram2025}. These approaches provide effective mechanisms for identifying informative visual elements, but communication introduces a different evaluation target: the usefulness of a transmitted token is ultimately determined by the quality of the image reconstructed after the complete packet has been received. A locally important token may contribute differently when combined with other transmitted tokens or when the receiver can infer its content from contextual information. Token selection for generative communication therefore benefits from a terminal-value criterion that captures both sequential token interactions and receiver-side completion.

Full-budget counterfactual evaluation provides such a criterion by evaluating each candidate action through the reconstruction obtained after completing the remaining transmission budget. GCR-C \cite{gcrc} implements this principle by comparing candidate tokens under the same Local-MDL continuation, packet budget, and receiver. Each resulting value measures the contribution of a candidate within a complete transmission sequence and therefore directly reflects terminal reconstruction quality. This evaluation also introduces a substantial encoder-side computational workload: every candidate query requires repeated prior inference, continuation of the remaining transmission decisions, and final image reconstruction. As a result, the transmitter must allocate two coupled resources---the communication bits used to represent the image and the computation used to determine how those bits should be spent.

This computation-allocation problem is related to selective prediction and learning-to-defer, where an expensive expert is invoked only for inputs that benefit most from additional evaluation \cite{geifman2017selective,lakshminarayanan2017ensembles,mozannar2020defer,hemmer2023limited}. In generative image communication, however, selective evaluation operates at both the state and candidate levels. The transmitter must determine which communication states merit additional refinement and, within each selected state, which candidate tokens should receive exact terminal evaluation. The resulting controller must therefore combine an estimate of terminal reconstruction value with the computational cost associated with candidate queries.

To address this problem, we propose ACV-Gate, a selective amortization framework that learns terminal candidate rankings from full-budget counterfactual evaluations and allocates exact evaluations to the candidates and communication states where they are most informative. A set-aware student learns candidate rankings from terminal advantages and regrets produced by the full-budget evaluator, allowing it to select a direct action without exact candidate evaluation and to rank alternative actions for subsequent refinement. A bounded candidate screen then retains the Local-MDL action, the direct student action, and highly ranked alternatives for exact evaluation. Cost-based state allocation further controls when this refinement is applied, enabling the encoder to trade candidate-query workload for reconstruction quality under explicit computational budgets.

\noindent\textbf{Relation to the terminal evaluator:}
GCR-C~\cite{gcrc} serves as the fixed full-budget evaluator from which ACV-Gate learns terminal candidate preferences. ACV-Gate preserves this terminal-quality objective while amortizing its repeated use during token selection. The learned student provides an immediate candidate ranking, and selective refinement concentrates exact GCR-C evaluations on a bounded set of promising alternatives. This formulation turns terminal evaluation from a computation applied uniformly across candidates into a controllable encoder resource that can be allocated according to its expected reconstruction benefit.

The main contributions are as follows:
\begin{enumerate}
\item We formulate token selection in generative image communication under three coupled resources: serialized packet bits, a per-image cap on exact candidate evaluations, and an average candidate-query budget. This formulation makes encoder computation an explicit resource alongside communication rate.

\item We develop a set-aware student that learns terminal candidate rankings from full-budget advantages and regrets. The learned ranking supports direct token selection without exact candidate evaluation and provides an ordered candidate set for bounded refinement.

\item We develop a reference-preserving selective controller that combines bounded candidate screening with cost-based state allocation. Matched-budget comparisons separate the effects of terminal-value ranking, candidate screening, and state allocation, while synchronized runtime measurements and evaluations across image systems characterize the resulting computation--quality trade-off.
\end{enumerate}

\section{Related Work}
\subsection{Semantic communication and learned image transmission}
Learned image compression optimizes analysis and synthesis transforms
together with entropy models
\cite{balle2016endtoend,balle2018hyperprior,minnen2018joint}. Vector
quantization and masked generation represent images as discrete tokens whose
missing entries can be predicted from context
\cite{oord2017vqvae,esser2021taming,chang2022maskgit}. Deep joint
source--channel coding learns image transmission over a noisy link, while
semantic and generative communication exploit task structure and receiver
priors
\cite{bourtsoulatze2019deep,xie2021deep,han2022generative}.

Recent communication methods address progressive transmission, content
adaptation, semantic masking and sparse visual representations
\cite{plit2025,catjscc2025,sqgan2025,sparsesbc2025}. Their design choices
concern the representation, channel mapping and rate allocation.
For a fixed representation and receiver, token selection still requires the
encoder to estimate the effect of each transmitted position. ACV-Gate
allocates the computation used to make that decision.

\subsection{Visual token selection and adaptive computation}
DynamicViT, EViT, TokenLearner, A-ViT and Token Merging reduce visual
computation through learned importance, token reorganization, adaptive
selection or merging
\cite{rao2021dynamicvit,liang2022evit,ryoo2021tokenlearner,yin2022avit,
bolya2023tome}. More recent methods use diversity, visual cues,
saliency--coverage, graph structure and attention centrality
\cite{divprune2025,vispruner2025,scope2025,gprune2025,tram2025}.
SparseVLM, LLaVA-PruMerge and FitPrune extend efficient token reduction to
multimodal inference
\cite{sparsevlm2025,prumerge2025,fitprune2025}. Analysis of encoder-layer
information further shows why a pruning rule's effectiveness depends on
where and how token information is represented \cite{horizon2026}.

Attention, diversity and coverage provide inexpensive estimates of token
importance. For packetized reconstruction, these estimates must reflect how
a selected token affects subsequent transmission and receiver completion. We
evaluate adapted DivPrune, VisPruner and SCOPE rules with a common proposal
and receiver. ACV-Gate learns its ranking from completed-image quality under
that same communication process.

\subsection{Selective prediction and decision distillation}
Selective prediction equips a predictor with a reject option, and
learning-to-defer extends this idea to costly expert access
\cite{geifman2017selective,lakshminarayanan2017ensembles,angelopoulos2022crc,
mozannar2020defer,hemmer2023limited}. In visual-token communication, the
expert is itself a computation: GCR-C completes the packet under a fixed
continuation and reconstructs the image to obtain terminal candidate value
\cite{gcrc}.

These strands motivate selective computation based on terminal reconstruction
value. ACV-Gate adds an amortization and budgeting layer to the GCR-C
evaluator. The student learns terminal candidate rankings, and the controller
allocates exact evaluations to bounded candidate sets. This links token
selection and encoder computation under a common packet budget and receiver.

\section{System Model and Problem Formulation}
Let $x$ be an image and $z=(z_1,\ldots,z_N)$ its discrete visual-token
sequence.  A transmitted set $S$ is represented by a packet containing a mode
field, a position description, token payload, a cyclic redundancy check (CRC)
and optional forward error correction (FEC). Let
$B(x)$ denote the bit budget assigned to $x$ at a given operating rate.  With
packet function $\mathsf{bits}(S)$, the feasible candidate set is
\begin{equation}
 \mathcal F(S,B)=\{a\notin S:\mathsf{bits}(S\cup\{a\})\leq B\}.
 \label{eq:feasible}
\end{equation}
The receiver uses a masked prior $D_\theta$ to complete unknown positions and
produces $\hat x(S)$.  A communication state is
$s_t=(S_t,B_{\mathrm{rem},t},\mathbf f_t,c_t)$, where $\mathbf f_t$ contains recoverability
and local-score features and $c_t$ encodes the operating rate, unselected-token fraction and stage.
The experiments use a fixed channel configuration.  The Local-MDL
action is
\begin{equation}
 \AL(s_t)=\arg\max_{a\in\mathcal F(S_t,B)}
 -\log p_\theta(z_a\mid z_{S_t}).
 \label{eq:local}
\end{equation}
To evaluate candidate $a$, the transmitter appends it to $S_t$ and completes
the packet using Local-MDL. The resulting receiver PSNR defines
$\QB(a\mid s_t)$. This exact teacher uses the same continuation, packet
syntax and reconstruction as the communication system. Relative to Local,
the teacher advantage and within-state regret are
\begin{equation}
 \begin{aligned}
 \AB(a\mid s_t)&=\QB(a\mid s_t)-\QB(\AL\mid s_t),\\
 \Reg(a\mid s_t)&=\max_{b\in\mathcal P(s_t)}\AB(b\mid s_t)
                     -\AB(a\mid s_t).
 \end{aligned}
 \label{eq:advantage}
\end{equation}
Thus $\Reg(a\mid s_t)\geq0$ within the compact proposal, and $a^*$ denotes
its minimum-regret candidate. Advantage and regret supervise the student
during training.

The compact proposal is the deduplicated union of Top-3 Local, Top-3
selected-set importance and Top-2 coverage candidates.  We denote it by
$\mathcal P(s_t)$ and write $K_s=|\mathcal P(s_t)|\leq 8$.
The four-candidate screen introduced below is a subset of this proposal.
We refer to the full-proposal expert as Exact-Full throughout the paper.

For a policy $\pi$, let $C_{\max}$ be the per-image hard cap on exact
candidate evaluations and let $q$ be the average candidate-query budget.
The constrained objective is
\begin{equation}
 \begin{aligned}
 \max_{\pi}\;&\frac{1}{|\mathcal X|}\sum_{x\in\mathcal X}Q(\pi;x)\\
 \text{subject to }&\mathsf{bits}(S_\pi(x))\leq B(x),\quad\forall x\in\mathcal X,\\
 &N_{\mathrm{cf}}(x)\leq C_{\max},\quad\forall x\in\mathcal X,\\
 &\frac{1}{|\mathcal X|}\sum_{x\in\mathcal X}N_{\mathrm{cf}}(x)\leq q.
 \end{aligned}
 \label{eq:computeallocation}
\end{equation}
where $Q(\pi;x)$ is terminal receiver quality for image $x$, $N_{\mathrm{cf}}(x)$
counts candidate-level exact $Q_B(a\mid s)$ evaluations, $S_\pi(x)$ is the
terminal transmitted set, and $C_{\max}$ is the hard per-image cap.
Here $B(x)$ counts serialized communication bits, $C_{\max}$ bounds exact
candidate workload for one image, and $q$ controls average workload across
an image set. The controller maintains
$C_{\mathrm{rem},0}=C_{\max}$ and
decrements it by the number of candidates evaluated after each expert call.
Threshold calibration sets the average operating point, and each experiment
reports its realized mean $N_{\mathrm{cf}}$/image together with encoder runtime.

Online evaluation uses one eligible refinement decision per image, at the
initial state, where the available candidate budget equals $C_{\max}$. After this decision,
both the direct and refined branches complete the packet with Local-MDL.

For a fixed patch size $p$ and image dimensions $(H,W)$, we define the token
grid and the three scale-aware accounting quantities
\begin{equation}
 \begin{aligned}
 \Ntok(H,W)&=\left\lfloor\frac{H}{p}\right\rfloor
              \left\lfloor\frac{W}{p}\right\rfloor,\\
 c_{\mathrm{MP}}&=\frac{\Ncf}{HW/10^6},\qquad
 c_{\mathrm{tok}}=\frac{\Ncf}{\Ntok(H,W)} .
 \end{aligned}
 \label{eq:scaleaccounting}
\end{equation}
The absolute $\Ncf$/image measures candidate workload; $c_{\mathrm{MP}}$
and $c_{\mathrm{tok}}$ normalize this workload by image area and token count.
Measured runtime captures the cost of each evaluation on a given token grid.

\section{ACV-Gate}
\begin{figure*}[t]
\centering
\includegraphics[width=0.98\textwidth]{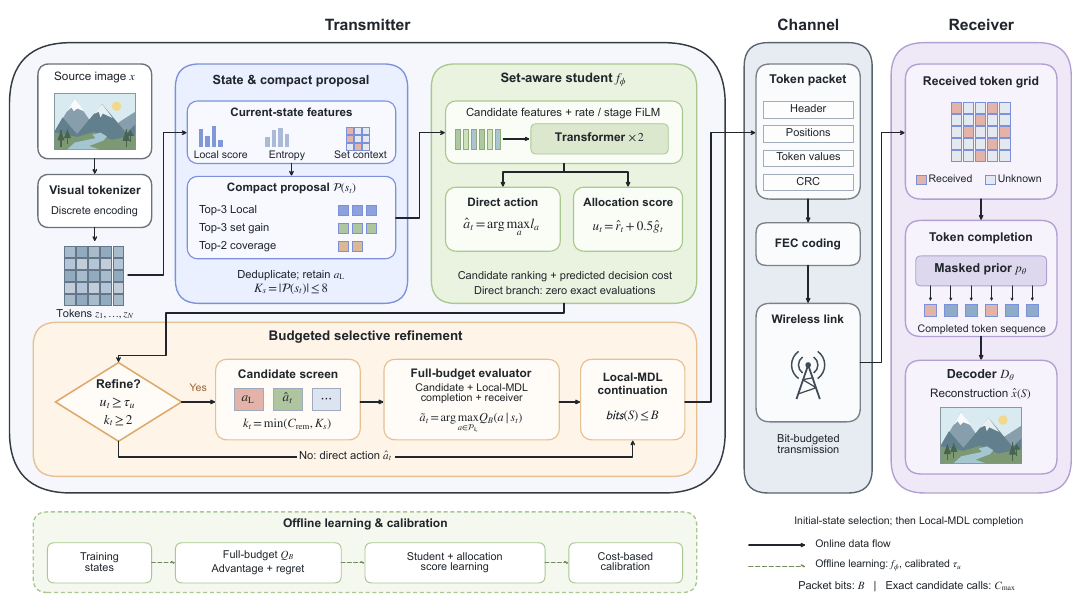}
\caption{ACV-Gate architecture. (a) Offline full-budget rollouts provide
candidate advantages and regrets for student and allocation-score learning.
(b) At the initial communication state, the student ranks the compact
proposal. The allocation score and remaining candidate budget select either
the direct action or a screened exact evaluation. (c) The selected action is
followed by Local-MDL continuation, packet transmission and receiver
reconstruction. Dashed arrows indicate transfer of learned parameters and
the calibrated threshold.}
\label{fig:framework}
\end{figure*}

\subsection{Fixed full-budget evaluator and compact proposal}
At training time, the full-budget evaluator labels the compact proposal
formed by the union of Top-3 Local, Top-3 selected-set importance
and Top-2 coverage candidates.  Duplicates are removed and Local is retained.
Each label is obtained by appending a candidate, completing the packet with
Local-MDL, and measuring the reconstructed image's PSNR. The Exact-Full
$Q_B$ reference evaluates the entire proposal with this same procedure
\cite{gcrc}. It provides the candidate optimum for the fixed proposal and
continuation.

\subsection{State and candidate representation}
Each candidate representation combines the masked-prior hidden state,
codebook vector, position, normalized remaining budget, Local score, entropy,
packet feasibility and proposal source. Relative features describe its
relationship to the proposal and selected tokens. These include standardized
scores, rank offsets, codebook and hidden-state distances, coverage and
redundancy. Source--rate interactions and normalized position distances
complete the set context. All input statistics are computed from the current
communication state; terminal advantage and regret provide the training
targets.

\subsection{Set-aware student and learning objective}
The candidate embeddings are processed jointly by a two-layer Transformer
without positional embeddings.  Hence permuting the proposal rows permutes
the output scores while preserving the set-level context.  For a padded
candidate group, the student produces logits $l_a$ and two auxiliary cost
estimates $(\hat r_t,\hat g_t)$:
\begin{equation}
 (\{l_a\}_{a\in\mathcal P(s_t)},\hat r_t,\hat g_t)=f_\phi(\{\phi(s_t,a)\}_{a\in\mathcal P(s_t)},c_t).
 \label{eq:student}
\end{equation}
The direct action is $\hat a=\arg\max_a l_a$. The auxiliary outputs estimate
its regret and lost positive gain, giving the allocation score
$u_t^{\mathrm{ACV}}=\hat r_t+0.5\hat g_t$. We denote the controller's state
score by $u_t=u(s_t)$ and compare this learned score with student margin,
Local uncertainty and random allocation under a common candidate budget.
The condition $c_t$ contains a seven-way rate ID, normalized rate $r/\max r$,
unselected-token fraction $1-|S_t|/N$, and initial/early/mid stage ID. A small
feature-wise linear modulation (FiLM) network conditions each candidate
embedding on these variables.

\paragraph{Regret and gain supervision.}
Let $q_{ta}\propto\exp[-\Reg(a\mid s_t)/T_t]$ be the regret-soft teacher target
with $T_t=0.10$, and let $p_{ta}=\operatorname{softmax}(l_{ta}/T_s)$ with
$T_s=0.35$. The hard-target weight is
$\alpha_t=\sigma[(g_t-0.20)/0.06]$, where $g_t$ is the gap between the
teacher's two highest candidate values in dB.
The complete training objective is
\begin{equation}
 \begin{aligned}
 \mathcal L_{\mathrm{cls}}&=\alpha_t\operatorname{CE}(a^*,p)
 +(1-\alpha_t)\mathcal L_{\mathrm{soft}},\\
 \mathcal L_{\mathrm{H2}}&=\mathcal L_{\mathrm{cls}}
 +0.33\mathcal L_{\mathrm{reg}}+0.40\mathcal L_{\mathrm{pair}},\\
 \mathcal L_{\mathrm{ACV}}&=\mathcal L_{\mathrm{H2}}
 +0.10\mathcal L_{\mathrm{gain}}+0.05\mathcal L_{\mathrm{safe}} .
 \end{aligned}
 \label{eq:acvloss}
\end{equation}
Expected regret penalizes the terminal quality lost by the student, while
pairwise ranking gives larger penalties to costly action errors. With
$r_{ta}=[\max_b\AB(b\mid s_t)-\AB(a\mid s_t)]_+$,
$q_{ta}=\exp(-r_{ta}/0.10)/\sum_b\exp(-r_{tb}/0.10)$, and
$p_{ta}=\operatorname{softmax}(l_{ta}/0.35)$,
\begin{equation}
 \begin{aligned}
 \mathcal L_{\mathrm{soft}}&=-\sum_a q_{ta}\log p_{ta},\\
 \mathcal L_{\mathrm{reg}}&=\sum_a p_{ta} r_{ta},\\
 \mathcal L_{\mathrm{pair}}&=\operatorname{mean}_{t,\;a\in\mathcal P_t\setminus
 \{a_t^*\}} w_{ta}\,\operatorname{softplus}\!\left(
 m_{ta}-(l_{ta^*}-l_{ta})\right),\\
 w_{ta}&=0.25+0.75\min(r_{ta}/0.50,1),\\
 m_{ta}&=0.05+0.45\min(r_{ta}/0.50,1).
 \end{aligned}
 \label{eq:lossdetails}
\end{equation}
Let
$h_t=[\max_a\AB(a\mid s_t)]_+$,
$u_{ta}=[\AB(a\mid s_t)]_+/(h_t+10^{-6})$, and let $Q^+_{50}$ be the
training median of positive headrooms.  The gain and safety terms are
\begin{equation}
 \begin{aligned}
 \mathcal L_{\mathrm{gain}}&=\operatorname{mean}_{h_t>0.05}[\operatorname{clip}
 (h_t/Q^+_{50},0.5,2.0)\\
 &\qquad\times(1-\sum_a p_{ta}u_{ta})],\\
 \mathcal L_{\mathrm{safe}}&=\operatorname{mean}_t\sum_a
 p_{ta}\frac{[-\AB(a\mid s_t)]_+}{0.50}.
 \end{aligned}
 \label{eq:gain-safety}
\end{equation}
The gain term rewards recovery of available positive headroom, and the
safety term penalizes actions whose terminal quality falls below Local.
Anchored adaptation combines these terms with H2 and a $0.35$ KL penalty
relative to the frozen H2 action distribution.
All padded rows are masked before the softmax and loss reduction.  The pair
term is zero for a group with fewer than two active candidates, and the gain
term is zero when no active state has $h_t>0.05$.
The training stages, optimizers, masking and parameter-freezing rules appear in
Supplementary Table~S13.

Training uses a 10-epoch hard-CE/pair warm-up, 30 regret-decision epochs and a
10-epoch anchored adaptation phase that updates the conditional student
blocks and score head.  After the direct checkpoint is selected, the selector
is frozen, and a 15-epoch allocation-score phase fits the nonnegative regret
and lost-gain outputs. We denote this complete configuration by Full.

\subsection{Budgeted selective expert allocation}
The direct branch selects the highest-scoring candidate and completes the
packet with Local-MDL. For selective refinement, $u_t$ ranks communication
states and a screen bounds the candidates evaluated at each selected state.
Rate-specific thresholds set the frequency of refinement. The learned ACV
configuration uses $u_t^{\mathrm{ACV}}$; equal-budget experiments compare
it with student margin, Local uncertainty and random allocation.

Let $C_{\mathrm{rem},t}$ denote the remaining candidate-level budget for the
current image and let $\hat a_t=\arg\max_a l_a$ be the direct action. At a
selected state, the remaining budget determines the screen size. We first
form the priority list
$\mathcal L_t=[\AL(s_t),\hat a_t,
\operatorname{Rank}^{l}(\mathcal P(s_t)\setminus\{\AL(s_t),\hat a_t\})]$.
The screened proposal contains the first $k_t$ unique entries of this list:
\begin{equation}
 \begin{aligned}
 k_t&=\min\{C_{\mathrm{rem},t},|\mathcal P(s_t)|\},\\
 \mathcal P_{k_t}(s_t)&=\operatorname{FirstUnique}_{k_t}(\mathcal L_t).
 \end{aligned}
 \label{eq:screenedproposal}
\end{equation}
Here $\operatorname{FirstUnique}_{k}$ removes repeated positions in priority
order and stops after $k$ unique entries. The screen is filled even when Local
and the direct action coincide. Refinement requires at least two available
candidate slots. Define
$\tilde a_t^{(k)}=\arg\max_{a\in\mathcal P_{k_t}}\AB(a\mid s_t)$.  After an
expert evaluation, the budget state is updated by
$C_{\mathrm{rem},t+1}=C_{\mathrm{rem},t}-|\mathcal P_{k_t}(s_t)|$.
The online policy is therefore
\begin{equation}
 \pi_{\mathrm{adapt}}(s_t)=
 \begin{cases}
  \hat a_t,&u_t<\tau_u,\\
  \tilde a_t^{(k)},
  &u_t\geq\tau_u\ \wedge\ k_t\geq2,\\
  \hat a_t,&\text{otherwise.}
 \end{cases}
 \label{eq:adaptive}
\end{equation}
To relate these decisions to reconstruction quality, define the terminal
values of the Local action, direct action, full proposal and screened proposal:
\begin{equation}
 \begin{aligned}
 Q_{\mathrm L}(s_t)&=Q_B(\AL(s_t)\mid s_t),\\
 Q_{\mathrm D}(s_t)&=Q_B(\hat a_t\mid s_t),\\
 Q_{\mathrm E}(s_t)&=\max_{a\in\mathcal P(s_t)}Q_B(a\mid s_t),\\
 Q_k(s_t)&=\max_{a\in\mathcal P_{k_t}(s_t)}Q_B(a\mid s_t),\\
 H(s_t)&=Q_{\mathrm E}(s_t)-Q_{\mathrm L}(s_t),\\
 R_{\mathrm D}(s_t)&=Q_{\mathrm E}(s_t)-Q_{\mathrm D}(s_t),\\
 V_k(s_t)&=Q_k(s_t)-Q_{\mathrm D}(s_t).
 \end{aligned}
 \label{eq:value_decomposition}
\end{equation}
Here $H$ is the expert's headroom over Local, $R_{\mathrm D}$ is the direct
action's regret, and $V_k$ is the gain recoverable by the screen.
For $k_t<2$, the direct branch gives $Q_k=Q_{\mathrm D}$ and $V_k=0$.
If $\delta_t$ indicates that the expert branch is evaluated, the adaptive value
decomposes as
\begin{equation}
 \begin{aligned}
 Q_{\mathrm{Adaptive}}(s_t)&=Q_{\mathrm D}(s_t)+\delta_t V_k(s_t),\\
 Q_{\mathrm{Adaptive}}(s_t)-Q_{\mathrm L}(s_t)
 &=H(s_t)-R_{\mathrm D}(s_t)+\delta_t V_k(s_t).
 \end{aligned}
 \label{eq:adaptive_decomposition}
\end{equation}
Each called screen retains Local and the direct action under the evaluator's
deterministic continuation. It therefore satisfies
$Q_k(s_t)\geq\max\{Q_{\mathrm D}(s_t),Q_{\mathrm L}(s_t)\}$ and
$V_k(s_t)\geq0$. Equation~\eqref{eq:adaptive_decomposition} separates the
student's recovered gain, $H(s_t)-R_{\mathrm D}(s_t)$, from the additional
recoverable value $\delta_tV_k(s_t)$ supplied by exact screening.

Each screened evaluation adds $|\mathcal P_{k_t}|$ to
$N_{\mathrm{cf}}$, so the invariant $N_{\mathrm{cf}}(x)\leq C_{\max}$
holds by construction.  When the cap is large enough, the endpoint evaluates
the full compact proposal. The fixed-screen experiments use $k_t=4$ for
every eligible call.

\subsection{Training and threshold calibration}
Training uses image-grouped fitting and monitor splits. The monitor selects
the direct checkpoint, which is then fixed during allocation-head fitting.
For the common-screen comparison, all four student families use student
margin for state allocation.

Cost-based calibration chooses a threshold using realized candidate
expenditure on $M$ monitor images. Let
$\mathcal T=\{+\infty\}\cup\{u_i:i\in\mathcal D_{\mathrm{cal}}\}$, where
$\mathcal D_{\mathrm{cal}}$ is the monitor image set.  Then
\begin{equation}
 \begin{aligned}
 b_{\mathrm{cal}}(\tau)&=\frac{1}{M}\sum_{i=1}^{M}k_i
 \mathbf 1\{u_i\geq\tau,\ k_i\geq2\},\\
 \tau^*(b_0)&\in\underset{\tau\in\mathcal T:\,b_{\mathrm{cal}}(\tau)\leq b_0}
 {\arg\max}\;b_{\mathrm{cal}}(\tau),
 \end{aligned}
 \label{eq:actualcostcalibration}
\end{equation}
where $k_i$ is the realized screen size.  Ties are resolved by choosing the
largest threshold. The $+\infty$ threshold provides a zero-call option,
and the selected point maximizes realized expenditure within $b_0$.
Applying the fixed threshold to validation yields the transferred average
workload $q_{\mathrm{test}}$, while the per-image cap remains enforced during
execution.

\section{Experimental Protocol}
\subsection{Datasets and comparison protocol}
The main CIFAR-10 study uses the frozen discrete tokenizer, masked-prior receiver,
codebook, compact proposal, packet syntax and reconstruction code from GCR-C.
It contains 200 development and 200 image-disjoint validation images,
seven rates $\{0.16,0.20,0.28,0.32,0.40,0.44,0.52\}$ and
$2\times200\times7\times2=5,600$ state groups across data roles, images,
rates and initial/early states. Development groups are used for parameter fitting,
checkpoint selection, allocation-head fitting and threshold calibration.
Validation groups provide the reported performance evaluation. The primary
end-to-end policies use three independently initialized selector seeds
($20260817$, $20260818$, and $20260819$) on the same 200 validation images.
The independent threshold-selection evaluation uses 100 Calibration and
400 disjoint Test images; measured-cost transfer uses a 40-image
development monitor and the 200-image validation split. The STL-10 and
high-resolution evaluations are described below. The supplement specifies
image assignments, training settings and the matched component comparisons.

Within each image system, all methods share the tokenizer, packet syntax,
CRC/FEC accounting, receiver and Local-MDL continuation. The primary
end-to-end comparison evaluates the full proposal when the adaptive branch
is selected ($k_t=K_s$, $C_{\max}=8$). The common-screen and allocation
comparisons use $C_{\max}=4$, retaining Local and the direct action before
adding student-ranked alternatives. A common screening rule therefore gives
each student its own four-candidate set. In the quota comparisons, the top
$\lfloor Mq/4\rfloor$ states receive a call; measured-cost transfer instead
applies a fixed calibration threshold. Larger student margins receive higher
priority in the margin control.

The unified timing experiment measures Local, Exact-Full, Local-ranked-4 and
Full/A1 direct and adaptive paths in one warmed CUDA process. A second
experiment repeats the Local and expert branches five times. Supplementary
Table~S14 defines their measurement boundaries and aggregation; Table~S15
summarizes the data roles. Image identity is the splitting and bootstrap unit.

The scale-transfer study uses a frozen LlamaGen VQ-16 tokenizer and decoder
at 384$\times$384, a 24$\times$24 token grid and a locally trained masked
prior. For the expert comparison, deterministic DIV2K crops train the prior,
and eight DIV2K validation crops calibrate the packet budgets. All 24 Kodak
center crops are evaluated at two operating points. Local-MDL and Exact-Full
share an initial-state proposal of Top-3 Local and Top-5 coverage candidates.
Each token uses 14 code bits; the packet has a 32-bit header, one
adaptive-min position description, a 16-bit CRC and a $1.25\times$ FEC factor.
The calibrated budgets are 2,101 and 3,804 bits for target fractions 0.15 and
0.30, respectively. The receiver completes the token sequence using the
frozen masked prior and LlamaGen decoder under error-free packet delivery.
Supplementary Table~S12 reports a separately trained 256-source student on
Internal32 and Tecnick40, with training and allocation settings in the
accompanying text.

\subsection{Metrics}
We report peak signal-to-noise ratio (PSNR), structural similarity (SSIM),
actual bits per pixel (bpp), mean teacher regret, retained positive gain,
catastrophic regret (regret $>0.5$~dB), candidate-level exact evaluations
$\Ncf$/image, $\Ncf$/MP, $\Ncf$/token, expert invocations/image, internal
$N_{\mathrm{prior}}$ and $N_{\mathrm{roll}}$ counters, and synchronized CUDA
runtime.  Gain
recovery and candidate-level query fraction are defined relative to the fixed
Exact-Full $Q_B$ reference as
\begin{equation}
 \mathrm{GR}(m)=\frac{Q_m-Q_{\mathrm{Local}}}
 {Q_{\mathrm{Exact}}-Q_{\mathrm{Local}}},\qquad
 \mathrm{QF}(m)=\frac{N_{\mathrm{cf}}(m)}{N_{\mathrm{cf}}(\mathrm{Exact})}.
 \label{eq:headline_metrics}
\end{equation}
An expert invocation sends one communication state to the exact branch and
can incur several candidate evaluations. Encoder runtime is measured with
synchronized CUDA boundaries and complements the candidate count
$N_{\mathrm{cf}}$. Across-seed results are reported as
means$\pm$standard deviations. Paired image-bootstrap intervals resample
image IDs while retaining all associated states and rates.
The primary quality results use three selector seeds; the unified runtime
comparison uses a fixed seed and paired image-bootstrap intervals.

\section{Results}
We first examine how the terminal objective changes candidate ranking and
end-to-end reconstruction quality. We then evaluate candidate screening,
budget allocation, measured runtime and transfer across image systems.

\subsection{Terminal horizon changes candidate ranking}
We use the same compact proposal, state and receiver to compare
an immediate one-step target $Q_{\mathrm{imm}}$ with the full-budget terminal
target $Q_B$. The immediate branch stops after adding the candidate, whereas
the terminal branch completes the remaining packet with Local-MDL. We report
the terminal quality of the immediate choice, terminal regret and action
agreement.

At 0.20~bpp, immediate-value selection achieves a terminal gain of
$+0.198$~dB over Local, compared with $+1.353$~dB for terminal-value
selection. The immediate choice incurs 1.155~dB terminal regret and agrees
with the terminal choice on 23.0\% of images. Accounting for the remaining
transmission therefore changes both the preferred action and its
reconstruction benefit.

\begin{figure}[t]
\centering
\includegraphics[width=\columnwidth]{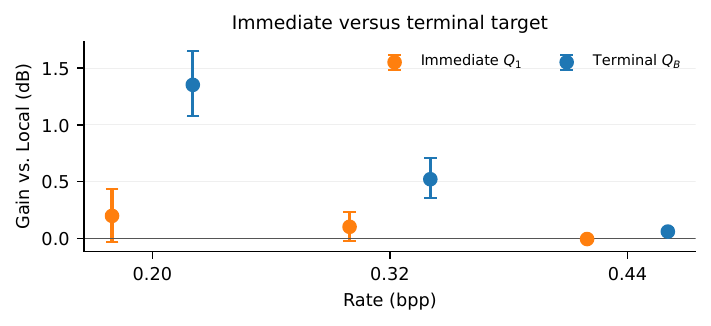}
\caption{Effect of evaluation horizon on candidate selection. Immediate and
full-budget selection use the same proposal, state and receiver. Both
selected actions are evaluated using the full-budget terminal objective.}
\label{fig:v5e5}
\end{figure}

\subsection{End-to-end reconstruction quality}
Table~\ref{tab:end2end} evaluates Local, Direct, Adaptive and Exact-Full $Q_B$
under the same packet and receiver settings. Three independently selected
Full checkpoints are evaluated on the same 200 validation images; Local and
Exact-Full are fixed references.

At 0.20~bpp, Direct gains $0.313\pm0.088$~dB with zero exact candidate
evaluations. In the primary full-proposal adaptive configuration, the gain
reaches $0.636\pm0.119$~dB with $2.13\pm0.53$ candidate evaluations per
image, recovering 46.99\% of the Exact-Full expert gain with 27.60\% of its
candidate calls. The synchronized runtime comparison in
Fig.~\ref{fig:v6frontier} evaluates Direct and the four-candidate Adaptive-4
configuration with fixed Full and A1 checkpoints.

Reconstruction gains decrease as the packet budget increases. At 0.32 and
0.44~bpp, Adaptive recovers
18.22\% and 12.18\% of the expert gain using 16.05\% and 8.14\% of its
calls. The expert's available gain itself falls from 1.353~dB at 0.20~bpp
to 0.060~dB at 0.44~bpp.

\begin{table*}[t]
\caption{End-to-end quality and candidate workload under identical packet
and receiver settings.  Direct and Adaptive ACV-Gate entries are means
$\pm$ standard deviations over three selector seeds on the same 200
image-disjoint validation images; Local and Exact-Full $Q_B$ are fixed 200-image
protocol references. Adaptive uses the full proposal on called images
($C_{\max}=8$). $N_{\mathrm{cf}}$ counts candidate-level exact calls.
Gain recovery (GR) and exact-query fraction (QF) use the corresponding fixed
Exact-Full $Q_B$ row as denominator. Actual bpp is 0.1986, 0.3156 and 0.4375
at the three rates.}
\label{tab:end2end}
\centering \tiny
\resizebox{\textwidth}{!}{%
\begin{tabular}{llrrrr}
\toprule
Rate & Method & $\Delta$Local (dB) & $N_{\mathrm{cf}}$/img. & GR (\%) & QF (\%)\\
\midrule
0.20 & Local & 0.000 & 0.00 & 0.00 & 0.00\\
     & Direct ACV-Gate & +0.313$\pm$0.088 & 0.00 & 23.15 & 0.00\\
     & Adaptive ACV-Gate & +0.636$\pm$0.119 & 2.13$\pm$0.53 & 46.99 & 27.60\\
     & Exact-Full $Q_B$ expert & +1.353 & 7.71 & 100.00 & 100.00\\
\midrule
0.32 & Local & 0.000 & 0.00 & 0.00 & 0.00\\
     & Direct ACV-Gate & +0.020$\pm$0.061 & 0.00 & 3.75 & 0.00\\
     & Adaptive ACV-Gate & +0.095$\pm$0.071 & 1.24$\pm$0.25 & 18.22 & 16.05\\
     & Exact-Full $Q_B$ expert & +0.521 & 7.71 & 100.00 & 100.00\\
\midrule
0.44 & Local & 0.000 & 0.00 & 0.00 & 0.00\\
     & Direct ACV-Gate & $-0.000$ $\pm$0.010 & 0.00 & $-0.13$ & 0.00\\
     & Adaptive ACV-Gate & +0.007$\pm$0.012 & 0.63$\pm$0.16 & 12.18 & 8.14\\
     & Exact-Full $Q_B$ expert & +0.060 & 7.71 & 100.00 & 100.00\\
\bottomrule
\end{tabular}
}
\end{table*}

The paired image intervals for the adaptive gain are positive at 0.20~bpp
and cross zero at 0.32 and 0.44~bpp. Low-rate communication therefore offers
the clearest benefit from allocating terminal reasoning, while higher rates
provide less reconstruction headroom.

Selective refinement also reduces the frequency of reconstruction losses
relative to Local.
The fraction of validation images with $\Delta$PSNR below $-0.1$~dB
is 31.5\% for Direct and 23.0\%
for Adaptive at 0.20~bpp; the corresponding fractions are 19.0\%/18.0\% at
0.32~bpp and 4.0\%/3.5\% at 0.44~bpp. The largest reduction occurs at
0.20~bpp, consistent with the larger mean benefit of refinement at low rate.

\subsection{Common-screen terminal-value comparison}
Four student families use the same four-candidate screening rule and
student-margin allocation. Their candidate rankings determine the direct
action and the alternatives in the screen. We evaluate these choices with
cached terminal $Q_B$ values. Each point averages 200 validation images and
three selector seeds.

At 0.20~bpp, Full, A1 Set-context, Row-valueMSE and DeepSets-regret-soft
achieve direct gains of 0.313, 0.430, 0.402 and 0.414~dB, respectively.
These direct decisions require zero exact candidate evaluations. At a target
of one candidate evaluation per image, the gains rise to 0.545, 0.595,
0.593 and 0.603~dB. At the four-call endpoint, they reach 1.119, 1.142,
1.077 and 1.166~dB. At 0.32~bpp, the one-call gains are 0.093, 0.128,
0.124 and 0.032~dB; at 0.44~bpp, all methods remain within 0.006~dB of
Local at this budget.

The four-call endpoint isolates candidate screening because every image
receives the same evaluation budget. At 0.20~bpp, learned screens achieve
1.077--1.166~dB across the four students, compared with 0.271~dB for
Local-ranked-4. Ranking candidates by learned terminal value thus retains
substantially more reconstruction gain within the same four exact evaluations.

\begin{figure*}[t]
\centering
\includegraphics[width=0.98\textwidth]{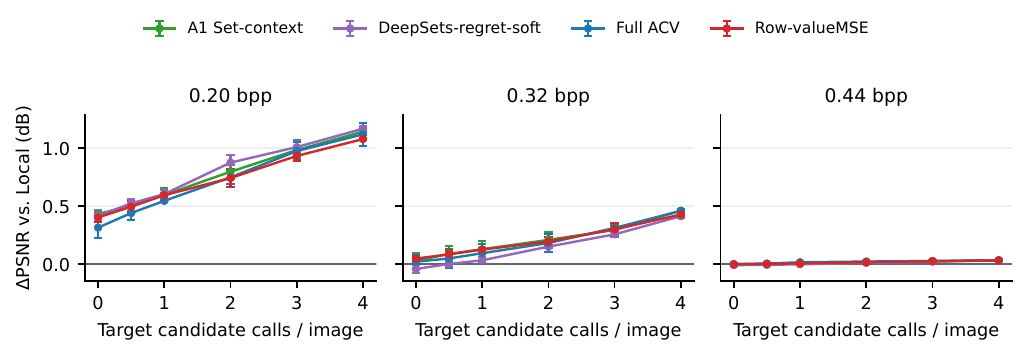}
\caption{Common-screen reconstruction gain versus candidate workload. The four
student families share a four-candidate screening rule and student-margin
allocation. Values
are mean $\Delta$PSNR relative to Local over three seeds; bars show seed
standard deviations. Terminal outcomes are evaluated from cached $Q_B$ values.}
\label{fig:v5e1frontier}
\end{figure*}

\begin{figure*}[t]
\centering
\includegraphics[width=0.98\textwidth]{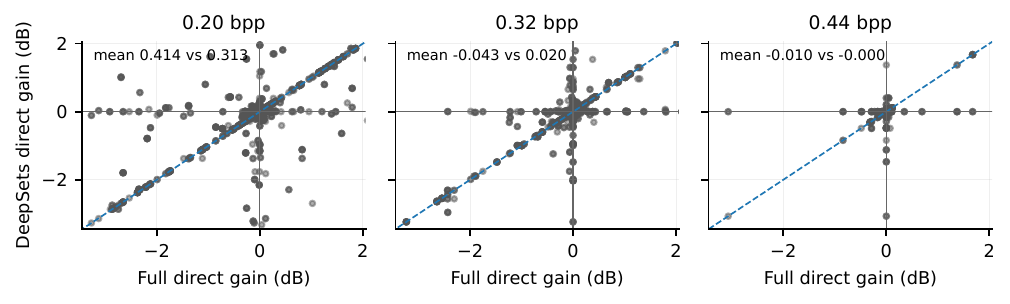}
\caption{Paired direct gains for Full and DeepSets-regret-soft on the same
validation images. The diagonal denotes equal gain, and the zero axes mark
the Local baseline. Annotations report the mean gains.}
\label{fig:v5pairs}
\end{figure*}

Figure~\ref{fig:v5pairs} compares Full and DeepSets-regret-soft on paired
images. DeepSets-regret-soft reaches 0.874~dB at two calls per image and
0.20~bpp, with the relative ranking changing across rates. The subsequent
allocation experiments use Full, which combines the set-aware student and
learned allocation head.

\subsection{Measured candidate expenditure and threshold transfer}
We calibrate a threshold for a target of one candidate evaluation per image
using 40 development images per rate, reserving every fifth development image.
Each selected state contributes its realized screen size to the calibration
cost. We then apply the fixed threshold to the 200 validation images. For
Full, the realized workloads are 0.71, 0.60 and 0.99 evaluations per image
at 0.20, 0.32 and 0.44~bpp, respectively. The corresponding paired gains
are 0.482, 0.072 and 0.014~dB. The low-rate interval excludes zero, and
every validation image satisfies the four-call cap.

\begin{figure}[t]
\centering
\includegraphics[width=\columnwidth]{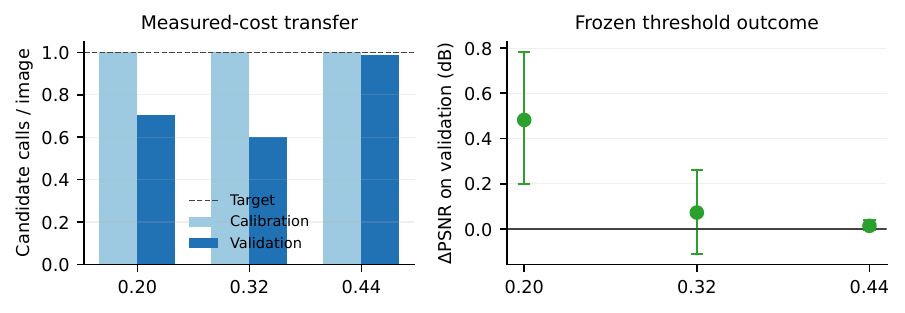}
\caption{Actual-cost threshold transfer for Full ACV-Gate. Thresholds are
calibrated on a reserved development monitor at a one-call target and frozen
before validation. Error bars are paired image-bootstrap intervals.}
\label{fig:v5e4}
\end{figure}

\subsection{Fixed-screen allocation-score controls}
\begin{figure*}[t]
\centering
\includegraphics[width=0.98\textwidth]{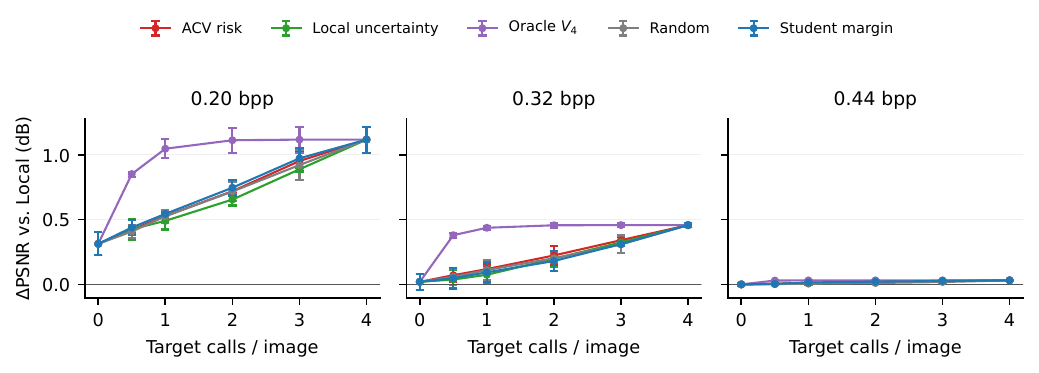}
\caption{Fixed-screen allocation-score controls.  The four practical scores
share the same Full direct student, four-candidate screen and target budgets;
the post-hoc oracle ranks by the realized $V_4$ screen value and gives the
allocation upper bound. Error bars denote seed standard deviations.}
\label{fig:v5e3}
\end{figure*}

To isolate state allocation, we fix the Full direct student and its
four-candidate screen, then vary the score used before exact evaluation.
We compare random, student-margin, Local-uncertainty and ACV-risk allocation
with an oracle that ranks states by the realized screen value $V_4$.
The oracle uses the same four candidates and measures the available
state-ranking headroom.
At 0.20~bpp and one candidate call per image, Random, Student margin, Local
uncertainty, and ACV risk achieve 0.526, 0.544, 0.491, and 0.528~dB,
respectively, while the same-screen $V_4$ oracle reaches 1.048~dB. These
matched-budget results place the practical allocation scores in a similar
operating range and identify state ranking as the main remaining source of
allocation headroom. At 0.32~bpp the
corresponding practical scores give 0.111, 0.093, 0.074 and 0.118~dB,
while the oracle gives 0.437~dB.  At 0.44~bpp all practical scores remain
at or below 0.015~dB at this budget, reflecting the small terminal headroom.
At the low-rate one-call budget, random allocation adds 0.213~dB to the
0.313~dB direct gain, and learned-risk allocation adds a further 0.002~dB.
Thus, screened refinement accounts for most of the observed gain at this
operating point, while the oracle gap quantifies the potential of improved
state ranking.
The complete allocation curves are shown in Fig.~\ref{fig:v5e3}; the
reported one-call values use the same validation images and seed aggregation
as the common-screen replay.

\subsection{Measured runtime--quality trade-off}
We measure Local-MDL, Exact-Full $Q_B$,
student-free Local-ranked-4, Full Direct, Full Adaptive-4, A1 Direct and A1
Adaptive-4 in one warmed CUDA process.  The same 200 validation images and
three operating rates are used throughout; the frozen Full and A1
checkpoints are the development-selected seed-$20260817$ instances.  The
student paths use one timed execution per image/rate. A separate branch-timing
experiment uses five repeated measurements and the per-image median.
All paths use batch size one and synchronized boundaries.  CPU tokenization is
measured once per image and added to each path; the student paths share one
measured Local/proposal prefix, whose duration is included in every student
row. Supplementary Table~S14 specifies the timing protocols.

\begin{figure*}[t]
\centering
\includegraphics[width=0.98\textwidth]{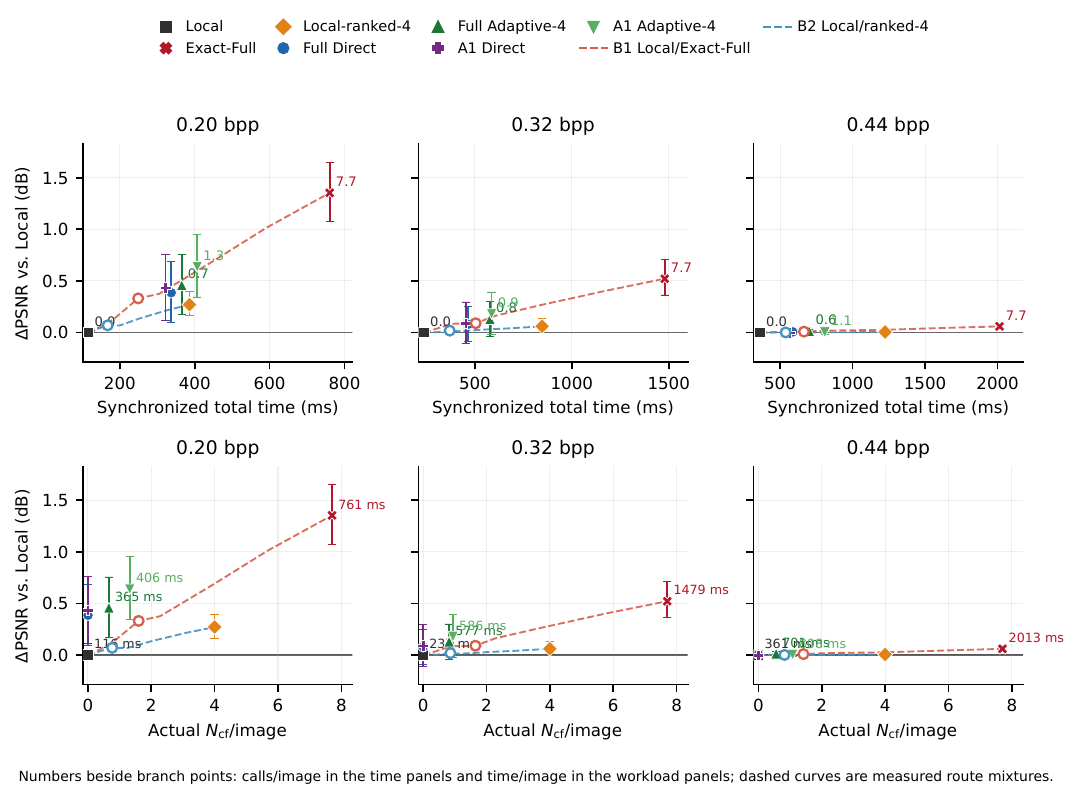}
\caption{Measured runtime--quality and workload--quality frontiers.
Upper panels use synchronized total time; lower panels use
realized candidate calls per image.  Error bars are paired image-bootstrap
intervals.  Solid markers are measured branches; dashed curves are B1
Local/Exact-Full and B2 Local/Local-ranked-4 route mixtures over the fixed route
seeds.  Full and A1 student paths use their frozen seed-$20260817$ instances;
the repeated five-pass branch measurements are given in Supplementary
Table~S4.}
\label{fig:v6frontier}
\end{figure*}

Figure~\ref{fig:v6frontier} reports the resulting matched measurements. The
upper panels plot paired PSNR gain against synchronized total runtime; the
lower panels plot the same gain against realized $\Ncf$/image. The
dashed curves mix Local with Exact-Full (B1) or Local-ranked-4 (B2)
using measured branch times and five fixed routing seeds.
Each operating point therefore specifies both the reconstruction gain and
its measured computational cost. The timing outputs also record the
95th-percentile per-image runtime.

In the unified run at 0.20~bpp, Full Direct achieves $+0.386$~dB at
337.0~ms with zero exact candidate evaluations. Full Adaptive-4 achieves
$+0.459$~dB at 365.1~ms with 0.66 evaluations per image. The corresponding
A1 gains are $+0.433$~dB at 322.0~ms and $+0.638$~dB at 406.4~ms;
A1 Adaptive-4 uses 1.32 evaluations per image. At 0.32~bpp, Full
Direct/Adaptive-4 gains are $+0.080$/$+0.130$~dB at 464.9/577.5~ms with
0/0.82 calls. A1 Direct/Adaptive-4 gains are $+0.087$/$+0.180$~dB at
452.5/585.9~ms with 0/0.94 calls.
At 0.44~bpp, all student gains are below $+0.013$~dB while the Exact-Full
reference remains $+0.060$~dB.

The repeated branch-timing experiment gives Local runtimes of
123.5, 264.0 and 387.2~ms at 0.20, 0.32 and 0.44~bpp, respectively. Exact-Full
gives $+1.353/+0.521/+0.060$~dB and 830.3/1587.2/2194.3~ms, while
Local-ranked-4 gives $+0.271/+0.061/+0.006$~dB and
416.1/900.8/1319.2~ms with four calls per image. At a routing probability
of $p=0.5$ and 0.20~bpp,
B1 obtains $+0.680$~dB at 482.8~ms with 3.94 calls per image, and B2 obtains
$+0.138$~dB at 273.8~ms with 2.05 calls. These operating points quantify
the extra computation required to improve on Local reconstruction.

\subsection{Comparison with adapted token-selection methods}
We adapt the diversity, visual-cue and saliency--coverage rules of DivPrune,
VisPruner and SCOPE from large vision--language models to packetized
reconstruction.
Each adapter selects from the ACV proposal using current-state features,
then completes the packet with Local-MDL. All methods use the same 200
validation images, tokenizer, receiver and packet accounting. Random-proposal
provides a proposal-level control, and the Exact-Full $Q_B$ expert
provides the terminal-value reference.

\begin{figure*}[t]
\centering
\includegraphics[width=0.98\textwidth]{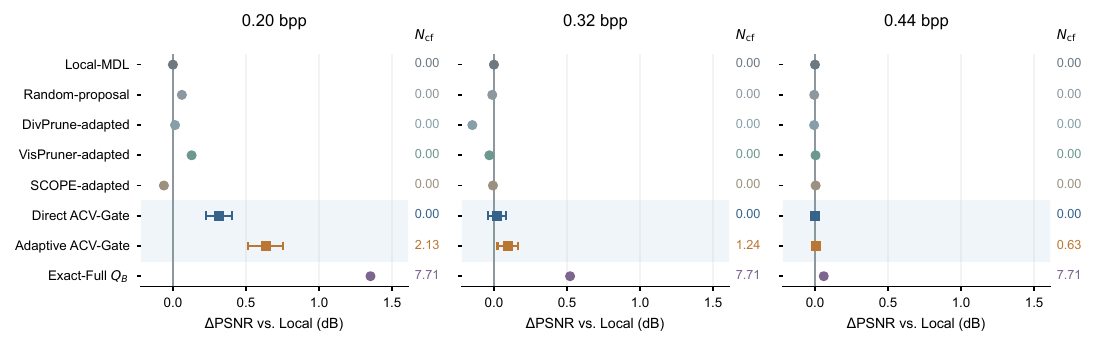}
\caption{Reconstruction gain of adapted token-selection methods and
ACV-Gate under the same packet and receiver. Direct and Adaptive are
three-seed Full means, with error bars denoting seed standard deviations.
The $N_{\mathrm{cf}}$ column in each panel reports mean candidate calls per
image.}
\label{fig:externalrules}
\end{figure*}

Figure~\ref{fig:externalrules} compares the reconstruction gains at each
candidate workload.
At 0.20~bpp, VisPruner-adapted improves over Local by 0.128~dB, while
DivPrune-adapted, SCOPE-adapted and Random-proposal change PSNR by +0.015, $-0.062$
and +0.061~dB, respectively.  At 0.32~bpp, all three structured adapters are
at or below Local and DivPrune-adapted is $-0.148$~dB.  At 0.44~bpp, all
adapters using zero exact candidate evaluations are within 0.006~dB of Local,
consistent with the small terminal headroom. The Exact-Full $Q_B$ expert gains
1.353, 0.521 and 0.060~dB
at the three rates with 7.705 candidate evaluations per image on average.
At 0.20~bpp, ACV-Gate Direct achieves +0.313$\pm$0.088~dB with zero exact
candidate evaluations, exceeding the gains of the three adapted rules. Adaptive reaches
+0.636$\pm$0.119~dB with $2.13\pm0.53$ calls. Terminal supervision thus
improves the low-rate direct decision, while selective evaluation recovers
additional quality at an intermediate candidate cost.

\subsection{Evaluation on STL-10}
We train the selector and allocation mechanism for a separate
96$\times$96 STL-10 tokenizer/receiver and evaluate them on 200 official
test images. Another 200 images form the development set, which is divided
into fitting and monitoring subsets. All three selector seeds are included
in the end-to-end evaluation. The budget is
$B=\operatorname{round}(r\times1024)$, where $r$ is the normalized budget
parameter shown on the horizontal axis. Actual bpp equals the serialized
packet length divided by $96^2$.

\begin{figure}[t]
\centering
\includegraphics[width=\columnwidth]{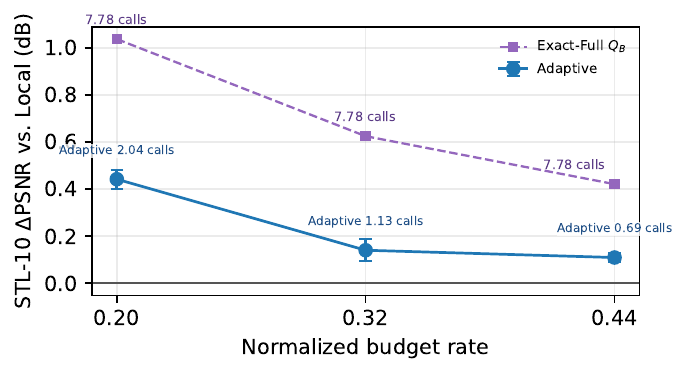}
\caption{Full adaptive ACV-Gate and the fixed Exact-Full $Q_B$ reference on the
independent STL-10 evaluation.  Adaptive points are three-seed means with seed
standard deviations; the dashed Exact-Full curve is annotated with its 7.78
candidate-level calls per image.  The reported paired image intervals for the
adaptive branch are positive at all three rates.}
\label{fig:stl10gain}
\end{figure}

Figure~\ref{fig:stl10gain} reports paired gains relative to the STL-10 Local
baseline. Full Adaptive gains $+0.442\pm0.041$, $+0.140\pm0.046$ and
$+0.110\pm0.020$~dB at $r=0.20$, $0.32$ and $0.44$, respectively.
The paired image intervals exclude zero at all three operating points.
Exact-Full achieves larger gains with 7.78 candidate evaluations per image.
This separately trained system reproduces the gain from selective computation
and its dependence on the transmission budget.

\subsection{High-resolution terminal-value learning and selective computation}
The Kodak-24 evaluation measures the reconstruction benefit and cost of
terminal evaluation on a 24$\times$24 token grid at 384$\times$384.
The initial-state Exact-Full expert improves
PSNR over Local by 1.033~dB at 0.0142~bpp and 0.847~dB at approximately
0.0257~bpp, using eight candidate calls per image. Its measured selection
runtime is 8.9 and 8.4 times that of Local, respectively (Supplementary
Fig.~S3).

At this resolution, a separately trained 256-source student achieves
low-rate adaptive gains of 0.553~dB on Internal32 and 0.520~dB on Tecnick40.
The corresponding workloads are 0.90 and 0.98 candidate evaluations per
image (Supplementary Table~S12). Both low-rate adaptive image-bootstrap
intervals exclude zero; higher-rate gains decrease and vary by dataset.
Internal32 tests generalization to sources held out from student training,
and Tecnick40 provides an external source distribution. The supplement
reports the five-seed results, allocation settings and source assignments
for prior and student training.

\section{Discussion}
Terminal reconstruction value links token selection to encoder computation.
The student ranks candidates with zero exact evaluations, and selective
refinement evaluates a compact set of alternatives.
Equation~\eqref{eq:adaptive_decomposition} separates the student's recovered
gain from the additional value obtained through screening.

The return on computation depends on the transmission budget. At low rates,
the receiver generates more content, so a selected token can substantially
affect completion. Larger budgets leave less Exact-Full headroom over Local,
reducing the benefit of additional evaluation. The separately trained STL-10
and $384\times384$ systems likewise concentrate their gains at low rates.

Compact students retain useful terminal rankings: A1 Set-context and
DeepSets-regret-soft match or exceed Full at several common-screen operating
points. State allocation provides further headroom. At 0.20~bpp and one
candidate evaluation per image, practical scores achieve 0.491--0.544~dB,
compared with 1.048~dB for the same-screen oracle. This gap identifies the
potential of learning when exact refinement is most valuable.

Deployment requires both workload and runtime measurements. The cap bounds
exact evaluations; runtime also includes feature extraction, student inference,
continuation and reconstruction. Their joint frontiers guide the choice of
computational budget. Extensions include budget reservation across
transmission states, perceptual terminal objectives and channel-conditioned
value estimates.

\section{Conclusion}
ACV-Gate combines terminal-value learning, bounded screening and budgeted
state allocation for visual token communication. On CIFAR-10 at 0.20~bpp, Direct
improves PSNR over Local-MDL by $0.313\pm0.088$~dB with zero exact candidate
evaluations. The primary full-proposal Adaptive configuration achieves
$0.636\pm0.119$~dB using $2.13\pm0.53$ evaluations per image, or 27.60\%
of the Exact-Full workload. Matched four-candidate experiments identify
terminal ranking and screening as the main sources of gain; allocation
controls quantify further headroom in state ranking. Separately trained
STL-10 and $384\times384$ systems also achieve their largest gains at low
rates. Measured runtime and candidate workload provide explicit operating
points for allocating encoder computation under a fixed packet budget.

\section*{Data and code availability}
The experiments use CIFAR-10, STL-10, DIV2K, Kodak and Tecnick images. The accompanying
source repository contains the selector, packet accounting, training and
evaluation scripts, with instructions for supplying datasets and model
artifacts.

\bibliographystyle{IEEEtran}

\begin{thebibliography}{40}
\scriptsize
\setlength{\itemsep}{-0.15ex}
\bibitem{balle2016endtoend}
J.~Ball\'e, V.~Laparra, and E.~P. Simoncelli, ``End-to-end optimized image
compression,'' in \emph{Proc. ICLR}, 2017.
\bibitem{balle2018hyperprior}
J.~Ball\'e, D.~Minnen, S.~Singh, S.~J. Hwang, and N.~Johnston, ``Variational
image compression with a scale hyperprior,'' in \emph{Proc. ICLR}, 2018.
\bibitem{minnen2018joint}
D.~Minnen, J.~Ball\'e, and G.~D. Toderici, ``Joint autoregressive and
hierarchical priors for learned image compression,'' in \emph{Proc. NeurIPS},
vol.~31, 2018.
\bibitem{oord2017vqvae}
A.~van den Oord, O.~Vinyals, and A.~Kavukcuoglu, ``Neural discrete
representation learning,'' in \emph{Proc. NeurIPS}, vol.~30, 2017.
\bibitem{esser2021taming}
P.~Esser, R.~Rombach, and O.~Ommer, ``Taming transformers for high-resolution
image synthesis,'' in \emph{Proc. CVPR}, 2021, pp.~12873--12883.
\bibitem{chang2022maskgit}
H.~Chang, H.~Zhang, L.~Jiang, C.~Liu, and W.~T. Freeman, ``MaskGIT: Masked
generative image transformer,'' in \emph{Proc. CVPR}, 2022, pp.~11315--11325.
\bibitem{bourtsoulatze2019deep}
E.~Bourtsoulatze, D.~B. Kurka, and D.~Gunduz, ``Deep joint source-channel
coding for wireless image transmission,'' \emph{IEEE Trans. Cogn. Commun.
Netw.}, vol.~5, no.~3, pp.~567--579, 2019.
\bibitem{xie2021deep}
H.~Xie, Z.~Qin, G.~Y. Li, and B.-H. Juang, ``Deep learning enabled semantic
communication systems,'' \emph{IEEE Trans. Signal Process.}, vol.~69,
pp.~2663--2675, 2021.
\bibitem{han2022generative}
T.~Han, J.~Tang, Q.~Yang, Y.~Duan, Z.~Zhang, and Z.~Shi, ``Generative model
based highly efficient semantic communication approach for image
transmission,'' arXiv:2211.10287, 2022.
\bibitem{plit2025}
G.~Zhang, H.~Li, Y.~Cai, Q.~Hu, G.~Yu, and Z.~Qin, ``Progressive learned
image transmission for semantic communication using hierarchical VAE,''
\emph{IEEE Trans. Cogn. Commun. Netw.}, vol.~11, no.~6, pp.~3640--3654,
2025, doi: 10.1109/TCCN.2025.3546935.
\bibitem{catjscc2025}
Y.~Li, X.~Chen, X.~Deng, and J.~Gui, ``Content adaptive distributed joint
source-channel coding for image transmission with hyperprior,'' \emph{IEEE
Trans. Cogn. Commun. Netw.}, vol.~11, no.~1, pp.~105--117, 2025, doi:
10.1109/TCCN.2024.3438371.
\bibitem{sqgan2025}
F.~Pezone, S.~Barbarossa, and G.~Caire, ``SQ-GAN: Semantic image
communications using masked vector quantization,'' \emph{IEEE Trans. Cogn.
Commun. Netw.}, early access, 2025, doi: 10.1109/TCCN.2025.3620819.
\bibitem{sparsesbc2025}
S.~Tong, X.~Yu, R.~Li, K.~Lu, Z.~Zhao, and H.~Zhang, ``Alternate
learning-based SNR-adaptive sparse semantic visual transmission,'' \emph{IEEE
Trans. Wireless Commun.}, vol.~24, no.~2, pp.~1737--1752, Feb. 2025, doi:
10.1109/TWC.2024.3512652.
\bibitem{rao2021dynamicvit}
Y.~Rao, W.~Zhao, B.~Liu, J.~Lu, J.~Zhou, and C.-J. Hsieh, ``DynamicViT:
Efficient vision transformers with dynamic token sparsification,'' in
\emph{Proc. NeurIPS}, vol.~34, 2021.
\bibitem{liang2022evit}
Y.~Liang, C.~Ge, Z.~Tong, Z.~Song, J.~Wang, and P.~Xie, ``Not all patches are
what you need: Expediting vision transformers via token reorganizations,'' in
\emph{Proc. ICLR}, 2022.
\bibitem{ryoo2021tokenlearner}
M.~S. Ryoo, A.~Piergiovanni, A.~Arnab, M.~Dehghani, and A.~Angelova,
``TokenLearner: What can 8 learned tokens do for images and videos?'' in
\emph{Proc. NeurIPS}, vol.~34, 2021.
\bibitem{yin2022avit}
H.~Yin, A.~Vahdat, J.~M. Alvarez, A.~Mallya, A.~Kautz, and P.~Molchanov,
``A-ViT: Adaptive tokens for efficient vision transformer,'' in \emph{Proc.
CVPR}, 2022, pp.~10809--10818.
\bibitem{bolya2023tome}
D.~Bolya, C.-Y. Fu, X.~Dai, P.~Zhang, C.~Feichtenhofer, and J.~Hoffman,
``Token merging: Your ViT but faster,'' in \emph{Proc. ICLR}, 2023.
\bibitem{geifman2017selective}
Y.~Geifman and R.~El-Yaniv, ``Selective classification for deep neural
networks,'' in \emph{Proc. NeurIPS}, vol.~30, 2017.
\bibitem{lakshminarayanan2017ensembles}
B.~Lakshminarayanan, A.~Pritzel, and C.~Blundell, ``Simple and scalable
predictive uncertainty estimation using deep ensembles,'' in \emph{Proc.
NeurIPS}, vol.~30, 2017.
\bibitem{angelopoulos2022crc}
A.~N. Angelopoulos, S.~Bates, A.~Fisch, L.~Lei, and T.~Schuster, ``Conformal
risk control,'' arXiv:2208.02814, 2022.
\bibitem{mozannar2020defer}
H.~Mozannar and H.~Sontag, ``Consistent estimators for learning to defer to an
expert,'' in \emph{Proc. ICML}, 2020, pp.~7076--7087.
\bibitem{hemmer2023limited}
P.~Hemmer, L.~Thede, M.~V\"ossing, J.~Jakubik, and N.~K\"uhl, ``Learning to
defer with limited expert predictions,'' arXiv:2304.07306, 2023.
\bibitem{divprune2025}
S.~R. Alvar, G.~Singh, M.~Akbari, and Y.~Zhang, ``DivPrune: Diversity-based
visual token pruning for large multimodal models,'' in \emph{Proc. IEEE/CVF
Conf. Comput. Vis. Pattern Recognit. (CVPR)}, 2025, pp.~9392--9401.
\bibitem{vispruner2025}
Q.~Zhang, A.~Cheng, M.~Lu, R.~Zhang, Z.~Zhuo, J.~Cao, S.~Guo, Q.~She, and
S.~Zhang, ``Beyond text-visual attention: Exploiting visual cues for effective
token pruning in VLMs,'' in \emph{Proc. IEEE/CVF Int. Conf. Comput. Vis.
(ICCV)}, 2025, pp.~20857--20867.
\bibitem{scope2025}
J.~Deng, W.~Li, J.~T. Zhou, and Y.~He, ``SCOPE: Saliency-coverage oriented
token pruning for efficient multimodel LLMs,'' in \emph{Advances in Neural
Information Processing Systems}, vol.~38, 2025.
\bibitem{sparsevlm2025}
Y.~Zhang, C.-K. Fan, J.~Ma, W.~Zheng, T.~Huang, K.~Cheng, D.~A. Gudovskiy,
T.~Okuno, Y.~Nakata, K.~Keutzer, and S.~Zhang, ``SparseVLM: Visual token
sparsification for efficient vision-language model inference,'' in
\emph{Proc. Int. Conf. Mach. Learn. (ICML)}, vol.~267, 2025, pp.~74840--74857.
\bibitem{prumerge2025}
Y.~Shang, M.~Cai, B.~Xu, Y.~J. Lee, and Y.~Yan, ``LLaVA-PruMerge: Adaptive
token reduction for efficient large multimodal models,'' in \emph{Proc.
IEEE/CVF Int. Conf. Comput. Vis. (ICCV)}, 2025, pp.~22857--22867.
\bibitem{fitprune2025}
W.~Ye, Q.~Wu, W.~Lin, and Y.~Zhou, ``Fit and Prune: Fast and training-free
visual token pruning for multi-modal large language models,'' in \emph{Proc.
AAAI Conf. Artif. Intell.}, vol.~39, no.~21, 2025, pp.~22128--22136.
\bibitem{gprune2025}
Y.~Jiang, Q.~Wu, W.~Lin, W.~Yu, and Y.~Zhou, ``What kind of visual tokens do
we need? Training-free visual token pruning for multi-modal large language
models from the perspective of graph,'' in \emph{Proc. AAAI Conf. Artif.
Intell.}, vol.~39, no.~4, 2025, pp.~4075--4083.
\bibitem{tram2025}
M.~Marchetti, D.~Traini, D.~Ursino, and L.~Virgili, ``Efficient token pruning
in vision transformers using an attention-based multilayer network,''
\emph{Expert Syst. Appl.}, vol.~279, Art. no.~127449, 2025, doi:
10.1016/j.eswa.2025.127449.
\bibitem{horizon2026}
Y.~Wang, J.~Wu, Z.~Ni, L.~Yang, Y.~Liu, C.~Yang, Y.~Wen, L.~He, X.~Tang,
H.~Liu, and Y.~Zhou, ``When token pruning is worse than random: Understanding
visual token information in VLLMs,'' in \emph{Proc. IEEE/CVF Conf. Comput.
Vis. Pattern Recognit. (CVPR)}, 2026, pp.~31910--31919.
\bibitem{gcrc}
 J.~Guo \emph{et al.}, ``Baseline-relative counterfactual refinement for
 bit-aware visual token communication,'' arXiv:2608.16192, 2026,
 doi: 10.48550/arXiv.2608.16192.
\end{thebibliography}

\end{document}